\pdfoutput=1
\documentclass[11pt]{article}

\usepackage[preprint]{acl}

\usepackage{times}
\usepackage{latexsym}

\usepackage[T1]{fontenc}
\usepackage[utf8]{inputenc}

\usepackage{microtype}

\usepackage{inconsolata}

\usepackage{amsmath}
\usepackage{mathtools}
\usepackage{amsfonts}
\usepackage{bm}
\usepackage{bbm}
\usepackage{booktabs}
\usepackage{tabularx}
\usepackage{tablefootnote}
\usepackage{float}
\DeclareMathOperator{\argmax}{argmax}
\DeclareMathOperator{\argmin}{argmin}
\DeclareMathOperator{\expect}{\mathbb{E}}

\title{Centroid-Based Efficient Minimum Bayes Risk Decoding}

\author{
Hiroyuki Deguchi${}^{1, 2}$ \
Yusuke Sakai${}^{1}$ \
Hidetaka Kamigaito${}^{1}$ \
Taro Watanabe${}^{1}$ \\
\textbf{Hideki Tanaka${}^{2}$ \
Masao Utiyama${}^{2}$} \\
${}^{1}$Nara Institute of Science and Technology \\
${}^{2}$National Institute of Information and Communications Technology \\
\texttt{\{deguchi.hiroyuki.db0, sakai.yusuke.sr9, kamigaito.h, taro\}@is.naist.jp} \\
\texttt{\{hideki.tanaka, mutiyama\}@nict.go.jp}
}

\begin{document}
\maketitle
\begin{abstract}

Minimum Bayes risk (MBR) decoding has achieved state-of-the-art translation performance using COMET, which is a neural metric that has a high correlation with human evaluation.
However, MBR decoding requires quadratic time because it computes the expected score between a translation hypothesis and all reference translations.
We propose centroid-based MBR (CBMBR) decoding to improve the speed of MBR decoding.
Our method clusters reference translations in the feature space and then calculates the score using the centroids of each cluster.
The experimental results demonstrate that our CBMBR not only improved the decoding speed of the expected score calculation by 5.7 times but also outperformed vanilla MBR decoding in terms of translation quality by up to 0.5 COMET\% in the WMT'22 En$\leftrightarrow$Ja, En$\leftrightarrow$De,  En$\leftrightarrow$Zh, and WMT'23 En$\leftrightarrow$Ja translation tasks.\footnote{\url{https://github.com/naist-nlp/mbrs}}

\end{abstract}

\section{Introduction}
Minimum Bayes risk (MBR) decoding achieves robust and high-quality translation by selecting the output sentence that maximizes the expected metric score
computed from the set of translation hypotheses~\citep{kumar-byrne-2004-minimum,eikema-aziz-2020-map,muller-sennrich-2021-understanding}.
Recently, neural evaluation metrics that have a high correlation with human evaluation have been proposed~\citep{rei-etal-2020-comet,rei-etal-2022-comet,sellam-etal-2020-bleurt,zhang-etal-2020-bertscore}, and MBR decoding using such neural metrics has achieved state-of-the-art translation performance in human evaluation compared with conventional maximum-a-posteriori (MAP) decoding using beam search~\citep{fernandes-etal-2022-quality}.

However, because of its formulation,
typical MBR decoding that regards the hypothesis set as a pseudo-reference set requires the computational time of $\mathcal{O}(N^2)$ when $N$ translation hypotheses are given.
In recent work, the number of hypotheses $N$ exceeded 1,000 candidates~\citep{freitag-etal-2023-epsilon}, which makes the quadratic order of computational time a challenge for MBR decoding, particularly when expensive neural metrics are used.
% To improve the decoding speed,
Several pruning methods have been proposed~\citep{eikema-aziz-2022-sampling,cheng-vlachos-2023-faster} to improve the decoding speed.
These approaches require the careful selection of a proxy metric~\citep{eikema-aziz-2022-sampling}, or it is difficult to take advantage of computational parallelism because hypotheses are pruned iteratively~\citep{cheng-vlachos-2023-faster}.

\begin{figure}[t]
    \centering
    \includegraphics[width=\linewidth]{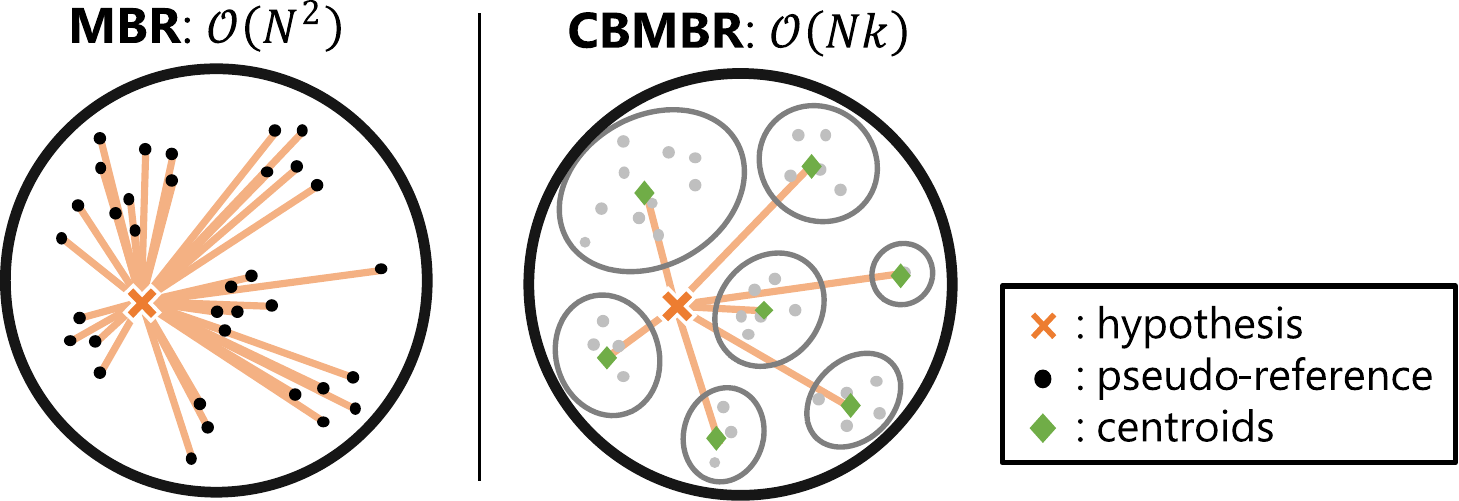}
    \caption{Overview of our centroid-based MBR (CBMBR).}
    \label{fig:CMBR-overview}
\end{figure}

Given that expensive neural metrics, e.g., COMET or BLEURT, are trained to output high scores when a hypothesis sentence and reference translation are semantically similar, we hypothesize that the distance between sentence vectors of similar sentences in the feature space of their models is close.
We leverage sentence similarity to improve the decoding speed of COMET-MBR by clustering sentence vectors of the translation candidates into $k \ll N$ clusters as shown in Figure~\ref{fig:CMBR-overview}.
Then, we calculate the COMET scores using $k$ centroid vectors of their clusters, instead of using $N$ sentence vectors.

In experiments, our proposed method not only achieved a speed-up of 5.7 times in the calculation of the expected score but also an improvement in the COMET score of up to 0.5\% compared with naive MBR decoding in the WMT'22 En$\leftrightarrow$Ja, En$\leftrightarrow$De, En$\leftrightarrow$Zh, and WMT'23 En$\leftrightarrow$Ja translation tasks.

\section{Background}
\paragraph{MBR decoding}
MBR decoding has been demonstrated to be effective in fields such as statistical automatic speech recognition~\citep{goel-and-byrne-2000-minimum} and statistical machine translation~\cite{kumar-byrne-2004-minimum}.
In recent years, it has been applied to neural machine translation~\citep{eikema-aziz-2020-map,muller-sennrich-2021-understanding}.
Furthermore, it is more suitable for multiple translation systems than ensemble models~\citep{ito-etal-2023-investigating}.

Let $\mathcal{X}$ and $\mathcal{Y}$ be the spaces of possible source sentences and target sentences, respectively.
MAP decoding generates the target sentence
$y^\ast_\text{MAP} \in \mathcal{Y}$ 
using $y^\ast_\text{MAP} = \argmax_{y \in \mathcal{Y}} p_\theta(y | x)$, 
where $\theta$ denotes the parameter of the translation model that calculates the likelihood of an output sentence $y$ given an input sentence $x \in \mathcal{X}$.
Because it is infeasible to calculate probabilities for all possible $y \in \mathcal{Y}$, beam search is typically used to obtain the solution.

By contrast, MBR decoding determines the output sentence $y^\ast_\text{MBR} \in \mathcal{Y}$ by maximizing the expected utility as follows:
\begin{align}
    y^\ast_\text{MBR} &= \argmax_{h \in \mathcal{H}} \expect_{\hat{y} \sim P(y | x)} \left[ u(h, \hat{y}) \right], \\
    &\approx \argmax_{h \in \mathcal{H}} \expect_{\hat{y} \in \hat{\mathcal{Y}}} \left[ u(h, \hat{y}) \right], \label{eq:MBR}
\end{align}
where $u\colon \mathcal{Y} \times \mathcal{Y} \to \mathbb{R}$ denotes the utility function, which represents the preference relation and
$\mathcal{H} = \{ h_i \}_{i=1}^{|\mathcal{H}|} \subset \mathcal{Y}$ 
denotes the set of translation hypotheses.
$P(y|x)$ is the true probability of being translated from a given input sentence $x \in \mathcal{X}$ and is approximated using the sampled reference translations 
$\hat{\mathcal{Y}} = \{\hat{y}_i \}_{i=1}^{|\hat{\mathcal{Y}}|} \subset \mathcal{Y}$, as shown in Equation~\ref{eq:MBR}, because the true probability is unknown.
Typical MBR decoding considers the hypothesis set itself as the pseudo-reference set, i.e., $\hat{\mathcal{Y}} \coloneqq \mathcal{H}$.
Note that the time complexity is $\mathcal{O}(N^2)$, where $N \coloneqq |\mathcal{H}|$, which is time-consuming.

\paragraph{COMET-MBR}
COMET is an evaluation metric of translation quality that achieves a high correlation with human evaluation.
The COMET model consists of the XLM-RoBERTa (XLM-R)-based sentence encoder~\citep{conneau-etal-2020-unsupervised} and the output layer, and is trained to predict direct assessment scores~\citep{rei-etal-2020-comet,rei-etal-2022-comet}.
It first encodes the source sentence $x \in \mathcal{X}$, hypothesis sentence $h \in \mathcal{Y}$, and reference sentence $\hat{y} \in \mathcal{Y}$ into their $D$-dimensional sentence vectors independently.
Then the COMET score is computed from the triplet of sentence vectors in the output layer.
Let $f\colon \mathcal{X} \cup \mathcal{Y} \to \mathbb{R}^D$ be the function of sentence encoding and $s\colon \mathbb{R}^{D} \times \mathbb{R}^{D} \times \mathbb{R}^{D} \to \mathbb{R}$ be the output layer.
The COMET score is computed using $s( f(x), f(h), f(\hat{y}) )$.
MBR decoding with COMET (COMET-MBR) replaces the utility $u$ in Equation~\ref{eq:MBR} with the COMET score:
\begin{align}
    &y^\ast_\text{COMET-MBR} \nonumber \\
    &=\argmax_{h \in \mathcal{H}} \expect_{\hat{y} \in \hat{\mathcal{Y}}} \left[ s(f(x), f(h), f(\hat{y})) \right].
    \label{eq:COMET-MBR}
\end{align}

\section{Proposed Method}
\label{sec:proposal}
Our proposed \emph{centroid-based MBR (CBMBR)} approximates the expected utility using the centroids of similar sentence vectors.
CBMBR decodes by computing the expected utility according to the following procedures: sentence encoding, clustering, and calculating the expected utility.

\paragraph{Encoding}
First, we compute the sentence vector of the source $f(x) \in \mathbb{R}^D$, the hypotheses
$\{ f(h_i) \}_{i=1}^{|\mathcal{H}|} \subset \mathbb{R}^D$, 
and the pseudo-references
$\{ f(\hat{y}_i) \}_{i=1}^{|\hat{\mathcal{Y}}|} \subset \mathbb{R}^D$.

\paragraph{Clustering}
Next, we cluster the sentence vectors of the pseudo-references into $k \ll N$ clusters and obtain the centroid vectors of each cluster
$\mathcal{C} = \{ \bm{c}_i \}_{i=1}^k \subset \mathbb{R}^D$.
Here, we employ $k$means++~\cite{arthur-and-vassilvitskii-k-means++} to prevent the centroids from being biased.
$k$means++ selects the initial centroids so that the distances between each pair of centroids are farther according to the weights calculated from the distances between vectors.
We describe the details of the algorithm in Appendix \ref{sec:kmeans++}.
Then, we cluster the vectors
$\{ f(\hat{y}_i) \}_{i=1}^{|\hat{\mathcal{Y}}|}$ 
using the standard $k$means algorithm.
Specifically, we calculate the following steps iteratively:
1) assign a vector to its nearest neighbor centroid and 2) update the centroid using the vectors assigned to its cluster.

\paragraph{Expected utility}
Finally, we calculate the expected utility by replacing pseudo-reference vectors $f(\hat{y}) \in \mathbb{R}^D$ with centroids $\bm{c} \in \mathbb{R}^D$ in Equation~\ref{eq:COMET-MBR}:
\begin{equation}
    y^\ast_\text{CBMBR} = \argmax_{h \in \mathcal{H}} \expect_{\bm{c} \in \mathcal{C}} \left[ s(f(x), f(h), \bm{c}) \right].
    \label{eq:CBMBR}
\end{equation}

The conventional method requires $\mathcal{O}(N^2)$ of computational time to compute the expected utility for all hypotheses, whereas our CBMBR computes it in $\mathcal{O}(Nk)$.
Note that $k$ ($1 \leq k \leq N$) is a hyperparameter that balances the trade-off between the decoding speed and approximation accuracy.
Especially, when $k=1$, i.e., $\mathcal{C} = \{ \bm{c}_1 \}$, the centroid $\bm{c}_1$ can be calculated as the average of all pseudo-reference vectors, i.e., $\bm{c}_1 = \frac{1}{|\hat{\mathcal{Y}}|}\sum_{i=1}^{|\hat{\mathcal{Y}}|} f(\hat{y}_i)$, and the time complexity of CBMBR is $\mathcal{O}(N)$, which is equivalent to \citet{denero-etal-2009-fast} and \citet{vamvas-and-sennrich-2024-linear-time}.
Our proposed method approximates the expected utility using centroid representations, which implicitly assumes that the score function $s$ is approximately linear.
Specifically, when $k=1$, we approximate the expected utility $\expect_{\hat{y}\in\hat{\mathcal{Y}}} \left[ s(f(x), f(h), f(\hat{y})) \right]$ as $s(f(x), f(h), \expect_{\hat{y}\in\hat{\mathcal{Y}}} \left[ f(\hat{y}) \right])$ in our method by assuming that the score function $s$ is roughly linear.

\section{Experiments}
\paragraph{Setup}
We conducted translation experiments with two settings;
one used diversified translation candidates and the other simulated a more realistic scenario, multi-system translation.
We evaluated the translation quality using the COMET score, which is the same as the utility function of MBR decoding used in our experiments.
For comparison, we also performed translation candidate reranking using a quality estimation model \textsc{CometKiwi}~\citep{rei-etal-2022-cometkiwi} (QE)\footnote{
Unlike MBR decoding, which requires calculating multiple pairwise scores from a single sentence representation, the QE model computes a single score from a single representation.
Additionally, \textsc{CometKiwi} does not compute explicit sentence vectors that are independent between the source and translation sentences but directly estimates the score from the concatenated two sentences.
Therefore, the QE model does not cache the sentence vectors of both the source and hypotheses, like MBR decoding.
}, MBR decoding with confidence-based pruning (PruneMBR)~\citep{cheng-vlachos-2023-faster},
and evaluated the quality upper bound (Oracle), which selects the hypothesis with the best score according to COMET using reference translations.
We also compared CBMBR without $k$means++, where we randomly selected the initial centroids from the sample set (w/o $k$means++).
We used \textsc{Comet-22}~\citep{rei-etal-2022-comet} for the evaluation metric and utility function.
In MBR decoding, we treat the hypothesis set as the pseudo-reference set, i.e., $\hat{\mathcal{Y}} \coloneqq \mathcal{H}$.
We set the number of centroids to $k=64$.
The details of our setup are shown in Appendix~\ref{sec:detail-settings}.

\paragraph{Diverse translation candidates}

\begin{table}[t]
    \centering
    \small
    \tabcolsep 2.1pt
    \begin{tabular}{@{}lrrrrrrr@{}}
        \toprule
        Decoding & en-ja & ja-en & en-de & de-en & en-zh & zh-en & avg. \\
        \midrule
        MAP & 78.7 & 69.7 & 77.3 & 79.2 & 77.4 & 70.1 & 75.4 \\
        QE & 86.6 & 76.2 & 82.2 & 82.1 & 82.9 & 76.9 & 81.2 \\
        MBR & \textbf{87.9} & \textbf{76.6} & \textbf{84.0} & \textbf{83.0} & \textbf{84.2} & \textbf{77.3} & \textbf{82.2} \\
        PruneMBR & \textbf{87.9} & \underline{76.5} & \textbf{84.0} & \textbf{83.0} & \underline{84.1} & \textbf{77.3} & \underline{82.1} \\
        CBMBR & \textbf{87.9} & \textbf{76.6} & \underline{83.9} & \textbf{83.0} & \underline{84.1} & 77.1 & \underline{82.1} \\
        ~~w/o $k$means++ & \underline{87.8} & 76.4 & 83.8 & \underline{82.9} & 84.0 & \underline{77.2} & 82.0 \\ \midrule
        Oracle & 90.6 & 81.9 & 87.0 & 86.5 & 87.7 & 81.2 & 85.8 \\
        \bottomrule
    \end{tabular}
    \caption{Translation quality
    in the WMT'22 translation task with the setting of diverse translation candidates.
    The best scores are emphasized in bold font and the second-best scores are underlined for each language direction.
    }
    \label{tab:results-unbiased}
\end{table}

\begin{table}[t]
    \centering
    \small
    \tabcolsep 4pt
    \begin{tabular}{@{}lrrrr@{}}
        \toprule
        Step & QE & MBR & PruneMBR & CBMBR \\
        \midrule
        Encode/hypotheses & -- & 247.0 & 248.0 & 247.8 \\
        Encode/source & -- & 51.6 & 51.1 & 51.2 \\
        Rerank & 450.1 & -- & -- & -- \\
        Prune & -- & -- & 5.5 & -- \\
        $k$means++ & -- & -- & -- & 36.5 \\
        Utility function; $s$ & -- & 322.2 & 79.6 & 20.1 \\
        \midrule
        E2E & 450.1 & 633.1 & 384.7 & 356.8 \\
        \bottomrule
    \end{tabular}
    \caption{Average processing time per sentence (msec) in
    the WMT'22 translation task in the diverse translation candidates setting.
    Note that ``E2E'' measures the end-to-end time from the sentence encoding to the expected utility calculation including miscellaneous processes.
    }
    \label{tab:results-speed}
\end{table}

In this setting, we evaluated translation quality in six language directions: En$\leftrightarrow$Ja, En$\leftrightarrow$De, and En$\leftrightarrow$Zh in the WMT'22 translation task~\citep{kocmi-etal-2022-findings}.
We generated translation candidates using the pre-trained multilingual translation model, M2M100~\citep{fan-etal-2021-beyond}.
We used beam search with a beam size of 256 for MAP decoding, and generated 1,024 translations using epsilon sampling with $\epsilon=0.02$~\citep{freitag-etal-2023-epsilon} for MBR decoding.

Table~\ref{tab:results-unbiased} shows the translation quality of each decoding method.
From the average scores in the table, compared with MAP decoding, both MBR decoding and the proposed CBMBR decoding improved the COMET score by +6.8 and +6.7\%, respectively, and the gap between Oracle and both MBR and CBMBR narrowed.
The results also show that the difference between CBMBR and MBR was within 0.1\% using $k$means++ initialization.

Next, we compared the decoding time of each method as shown in Table~\ref{tab:results-speed}.
From the table, the total time of COMET-MBR (E2E) indicates that CBMBR was 1.8 times faster than vanilla MBR.
Specifically, in the computation of the expected utility, which required quadratic time, the speed increased by 5.7 times when including $k$means++, and by 16.0 times when comparing only the utility computation.
We confirmed that, compared with PruneMBR, the speed of the expected utility calculation improved by 1.4 times.
One reason for this improvement is that, unlike PruneMBR,
CBMBR computes the expected utility in a single transaction, 
which makes it easier to leverage the parallel computation capabilities of the GPU.

To summarize, CBMBR maintains translation quality comparable with naive MBR decoding while accelerating the computational time of the expected utility calculation by 5.7 times, including clustering.

\paragraph{Multi-system translation}
\begin{table}[t]
    \centering
    \small
    \begin{tabular}{@{}lrrrrr@{}}
        \toprule
        & \multicolumn{2}{c}{WMT'22} & \multicolumn{2}{c}{WMT'23} \\
        \cmidrule(lr){2-3} \cmidrule(lr){4-5}
        Decoding & en-ja & ja-en & en-ja & ja-en & avg. \\
        \midrule
        MAP & 86.4 & 80.9 & 83.5 & 80.4 & 82.8 \\
        QE & 89.8 & 82.6 & 87.6 & 82.3 & 85.6 \\
        MBR & \underline{90.5} & \textbf{84.1} & 88.7 & \underline{83.7} & 86.7 \\
        PruneMBR & 88.9 & \underline{82.8} & 86.6 & 82.2 & 85.1 \\
        CBMBR & \textbf{90.9} & \textbf{84.1} & \textbf{89.2} & \textbf{83.8} & \textbf{87.0} \\
        ~~w/o $k$means++ & \underline{90.5} & \textbf{84.1} & \underline{88.8} & \underline{83.7} & \underline{86.8} \\
        \midrule
        Oracle & 93.4 & 89.4 & 91.9 & 88.5 & 90.8 \\
        \bottomrule
    \end{tabular}
    \caption{Translation quality 
    in the multi-system translation setting.
    }
    \label{tab:results-biased}
\end{table}

We also evaluated the effectiveness of our CBMBR in the setting where translation candidates are generated from multiple translation systems.
In particular, we followed the practice in \citet{deguchi-etal-2023-naist}, in which 18 candidate sets with each set comprising the 50-best translations were generated from nine models and two decoding methods:
beam search and top-$p$ sampling ($p=0.7$) with a beam size of 50.
We evaluated the translation quality in two language directions: En$\leftrightarrow$Ja in the WMT'22 and WMT'23 translation tasks~\citep{kocmi-etal-2022-findings,kocmi-etal-2023-findings}.

Table~\ref{tab:results-biased} shows the results.
Unlike the diverse translation candidates setting, CBMBR improved the translation quality by up to 0.5\% compared with naive MBR.
Naive MBR calculates the expected utility using all samples equally, which is prone to translation bias when candidates have a multimodal distribution.
By contrast, CBMBR estimates the expected utility using only centroids;
therefore, it decodes robustly, even if the distribution is multimodal.
A detailed analysis is shown in Appendix~\ref{sec:analysis-multimodality}.

To summarize, we found that CBMBR not only improved the decoding speed but also improved translation quality compared with vanilla MBR when the translation was determined from the candidate sets generated from multiple translation systems.

\section{Discussion}
\subsection{Number of centroids $k$}

\begin{figure}
    \centering
    \includegraphics[width=\linewidth]{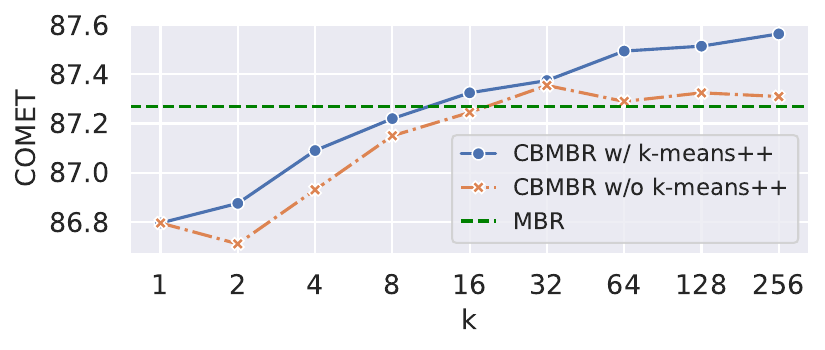}
    \caption{
    Translation quality of various $k$ in the multi-system translation setting.
    The scores are averaged COMET on WMT'22 En-Ja and Ja-En.
    }
    \label{fig:cbmbr-various-k}
\end{figure}

We evaluated the COMET scores of various $k \in \{ 2^i \}_{i=0}^{8}$ in the multi-system translation setting.
Figure \ref{fig:cbmbr-various-k} shows the results.
When $k=1$, while the time complexity was linear time $\mathcal{O}(N)$, the COMET score of CBMBR was degraded by 0.5\% compared with vanilla MBR.
The figure shows that translation quality improved as $k$ increased and CBMBR outperformed vanilla MBR when $k \geq 16$.
Additionally, translation quality was better when we used $k$means++ instead of the standard $k$means.

\subsection{Distance between similar sentence vectors}

\begin{table}[t]
    \centering
    \small
    \begin{tabular}{@{}lrr@{}} \toprule
        Model & dev & test \\ \midrule
        RoBERTa${}_\text{large}$~\cite{liu-etal-2019-roberta} & 53.7 & 43.0 \\
        fastText~\cite{joulin-etal-2017-bag} & 65.3 & 53.6 \\
        XLM-R${}_\text{large}$~\cite{conneau-etal-2020-unsupervised} & 39.1 & 31.6 \\
        LaBSE~\cite{feng-etal-2022-language} & 72.9 & 72.7 \\
        \midrule
        COMET~\cite{rei-etal-2022-comet} & 78.2 & 73.6\\
        \bottomrule
    \end{tabular}
    \caption{Pearson $r \times 100$ in the STS-B task using the encoder of the COMET model.}
    \label{tab:comet-sts}
\end{table}

As mentioned in Section~\ref{sec:proposal}, CBMBR implicitly assumes that the score function $s$ has linearity.
Because $s$ in the COMET model is a nonlinear multi-layer perceptron, it is not clear whether it is appropriate to use the averaged vector of hypothesis representations as representative points of the clusters.
To verify the assumption, we investigated the distances between sentence vectors of similar sentences using the semantic textual similarity benchmark (STS-B) task~\citep{cer-etal-2017-semeval}.
We evaluated the Pearson correlation coefficient $r$ using the ground truth similarity score.
Table~\ref{tab:comet-sts} shows the experimental results.
Despite sentence vectors not being trained explicitly like contrastive learning, COMET demonstrated a strong correlation of 73.6.
Moreover, the result demonstrates that it implicitly learned sentence similarity through the training of score prediction,
as evidenced by its significantly better correlation of 73.6 compared with the pre-trained XLM-R score of 31.6.
Furthermore, we confirmed that COMET outperformed LaBSE~\citep{feng-etal-2022-language} trained using contrastive learning.

To summarize, the sentence vectors of COMET demonstrated a strong correlation with gold scores in the STS-B task although the sentence representations were not explicitly trained.
Additionally, we confirmed that the encoder of COMET implicitly learned sentence similarity
through score prediction.
From the results, we verified that approximating the expected utility by aggregating sentence vectors that are close to each other is reasonable because the distance between sentence vectors represents semantic similarity.

\section{Conclusion}
In this paper, we proposed CBMBR, which improved the speed of MBR decoding by clustering the sentence vectors of similar sentences and computing the score using the centroid representations of each cluster.
Our CBMBR achieved a 5.7 times speed-up in the expected score calculation and an improvement in COMET of up to 0.5\% compared with vanilla MBR decoding in the WMT'22 En$\leftrightarrow$Ja, En$\leftrightarrow$De, En$\leftrightarrow$Zh, and WMT'23 En$\leftrightarrow$Ja translation tasks.
For future work, we will apply our method to other evaluation metrics, including both neural and non-neural metrics.

\section*{Limitations}
In this study focused only on improving the speed of MBR decoding, particularly the neural evaluation metric, COMET.
For non-neural metrics, it is necessary to apply the appropriate clustering method for each metric.

In COMET-MBR, there are two bottlenecks for the computational time:
the calculation of the expected utility and sentence encoding.
However, we only improved the computation speed of the expected utility, which took quadratic time.
Although sentence encoding can be computed in linear time, the sentences are encoded using the expensive XLM-R encoder, which is time-consuming.

Our method can only be applied to metrics for which we can compute the representation independently for each sentence.
This limitation is the same as that of the method of \citet{denero-etal-2009-fast} and is also explained in their paper.

We measured the decoding times reported in this paper on a single computer and only in a single run; the amount of speed improvement may differ when different computer architectures are used.

\section*{Ethical Consideration}
Both vanilla MBR decoding and CBMBR decoding select output sentences from a set of translation candidates generated by translation systems; hence, if the systems generate toxic text, it may be selected.

\section*{Acknowledgements}
This work was partially supported by JSPS KAKENHI Grant Number JP21H05054.

\bibliography{custom,anthology}
% \bibliography{custom}
\appendix

\section{Licenses}
In our experiments, we used the \textsc{Comet-22} model licensed under the 
Apache-2.0 license and \textsc{CometKiwi} model licensed under the CC BY-NC-SA 4.0 license.
We evaluated our method on the test sets of WMT'21, WMT'22, and WMT'23 translation tasks under the following policy: ``The data released for the WMT General MT task can be freely used for research purposes''.

\section{Details of the Datasets}

Table~\ref{tab:dataset-stats} shows the number of sentences for each dataset we used in our experiments.
\begin{table}[h]
    \centering
    \small
    \tabcolsep 3pt
    \begin{tabular}{@{}lrrrrrr@{}}
        \toprule
        Dataset & en-ja & ja-en & en-de & de-en & en-zh & zh-en \\
        \midrule
        WMT'21 & 1,000 & 1,005 & 1,002 & 1,000 & 1,002 & 1,948 \\
        WMT'22 & 2,037 & 2,008 & 2,037 & 1,984 & 2,037 & 1,875 \\
        WMT'23 & 2,074 & 1,992 & -- & -- & -- & -- \\
        \bottomrule
    \end{tabular}
    \caption{Number of sentences for each dataset we used.}
    \label{tab:dataset-stats}
\end{table}

\section{Details of the Algorithms and Models}
% \label{sec:appendix}

\subsection{$k$means++}
\label{sec:kmeans++}
We describe the algorithm for the initial centroid selection of $k$means++:
\begin{enumerate}
    \item Pick up the first centroid from the set $\hat{\mathcal{Y}}$ and add it to $\mathcal{C}$.
    \item Calculate the squared Euclidean distance between vector $f(\hat{y}_i)$ and its nearest centroid $d^2(\hat{y}_i) = \min_{\bm{c} \in \mathcal{C}} \lVert f(\hat{y}_i) - \bm{c} \rVert^2_2$. \label{enum:distance}
    \item Sample vector $f(\hat{y}_i)$ from the multinominal distribution according to the weights $\frac{d^2(\hat{y}_i)}{\sum_{j=1}^{|\hat{\mathcal{Y}}|} d^2(\hat{y}_j)}$ and add it to the set $\mathcal{C}$. \label{enum:sampling}
    \item Repeat steps~\ref{enum:distance} and~\ref{enum:sampling} until $k$ centroids are selected.
\end{enumerate}

\subsection{COMET model}
\begin{figure}[h]
    \centering
    \includegraphics[width=0.8\linewidth]{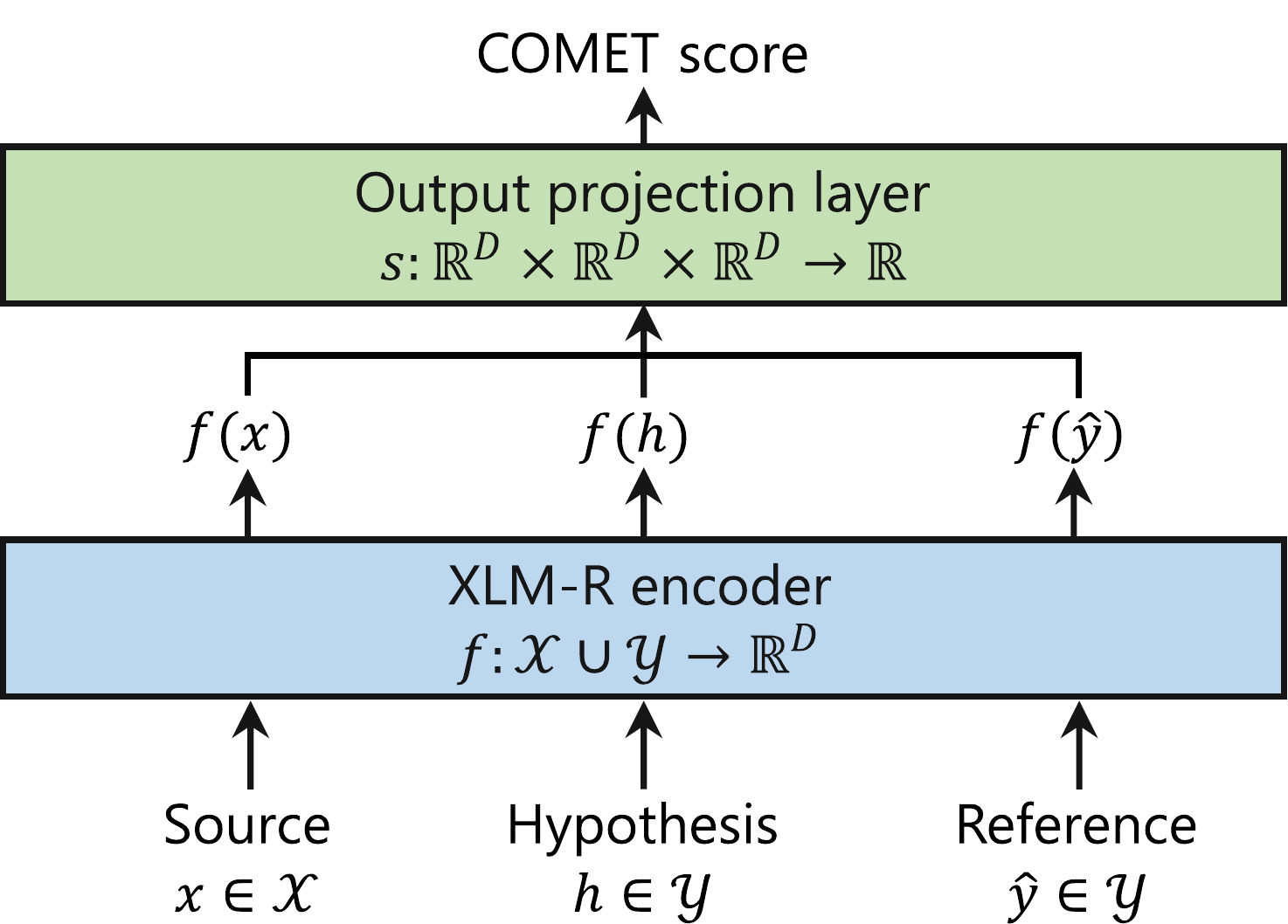}
    \caption{Overview of the COMET model.}
    \label{fig:model-comet}
\end{figure}

Figure~\ref{fig:model-comet} shows the overview of the COMET model.
A triplet of sentences are independently encoded into their sentence vectors, and then the COMET score is calculated from the vectors.

\section{Details of the Experimental Setup}
\label{sec:detail-settings}
Table~\ref{tab:detail-settings} shows the details of our experimental setup.
We implemented vanilla MBR, PruneMBR, and CBMBR using PyTorch.

\begin{table}[ht]
    \centering
    \small
    \tabcolsep 3pt
    \begin{tabular}{@{}ll@{}} \toprule
         Model \\
         ~~COMET & \texttt{Unbabel/wmt22-comet-da}\tablefootnote{\url{https://huggingface.co/Unbabel/wmt22-comet-da}} \\
         ~~QE & \texttt{Unbabel/wmt22-cometkiwi-da}\tablefootnote{\url{https://huggingface.co/Unbabel/wmt22-cometkiwi-da}} \\
         GPU & NVIDIA A100 $\times 1$ \\
         Batch size & 256 sentences \\
         (sentence encoding) & \\
         \midrule
         \multicolumn{2}{@{}l@{}}{\textbf{Diverse translation candidates setting}} \\
         Translation model & M2M100~\cite{fan-etal-2021-beyond} \\
         & (418M parameters)\tablefootnote{\url{https://huggingface.co/facebook/m2m100_418M}} \\
         MAP decoding & \\
         ~~~~Generation & beam search \\
         ~~~~Beam size & 256 \\
         MBR decoding \\
         \multicolumn{2}{@{}l@{}}{~~Candidate generation}  \\
         ~~~~\# of candidates & 1,024 translations \\
         ~~~~Generation & epsilon sampling ($\epsilon=0.02$) \\
         & \citep{freitag-etal-2023-epsilon} \\
         ~~CBMBR \\
         ~~~~\# of centroids $k$ & 64 \\
         ~~~~\# of iterations & 1 \\
         \midrule
         \multicolumn{2}{@{}l@{}}{\textbf{Multi-system translation setting}} \\
         Translation model & 9 various Transformer models \\
         & \citep{deguchi-etal-2023-naist} \\
         MAP decoding & \\
         ~~~~Generation & beam search using \\
         & the ensemble model \\
         ~~~~Beam size & 50 \\
         MBR decoding \\
         \multicolumn{2}{@{}l@{}}{~~Candidate generation} \\
         ~~~~\# of candidates & 900 translations\tablefootnote{\url{https://huggingface.co/naist-nlp/wmt23}} \\
         ~~~~Generation & beam search and \\
         & top-$p$ sampling ($p=0.7$) \\
         & \citep{deguchi-etal-2023-naist} \\
         ~~CBMBR \\
         ~~~~\# of centroids $k$ & 64 \\
         ~~~~\# of iterations & 1 \\
         
         \bottomrule
    \end{tabular}
    \caption{Details of our experimental setup.}
    \label{tab:detail-settings}
\end{table}

\section{Other Experimental Results}
\subsection{Translation quality on the development set in the diverse translation candidates setting}
Table~\ref{tab:results-mbr-valid} shows the experimental results of the diverse translation candidates setting on the development set.
In the table, ``niter'' denotes the number of iterations of $k$means clustering.
We chose niter = 1 from the results.

\begin{table}[h]
    \centering
    \small
    \tabcolsep 2.1pt
    \begin{tabular}{@{}lrrrrrrr@{}}
        \toprule
        Decoding & en-ja & ja-en & en-de & de-en & en-zh & zh-en & avg. \\
        \midrule
        MAP & 78.8 & 62.6 & 74.5 & 80.4 & 73.0 & 68.6 & 73.0 \\
        QE & 86.7 & 71.7 & 80.3 & 83.6 & 80.1 & 77.0 & 79.9 \\
        MBR & 88.2 & 72.6 & 82.1 & 84.3 & 81.6 & 77.7 & 81.1 \\
        PruneMBR & 88.2 & 72.6 & 82.0 & 84.3 & 81.6 & 77.7 & 81.1 \\
        CBMBR & 88.2 & 72.3 & 81.9 & 84.4 & 81.5 & 77.5 & 81.0 \\
        ~~w/o $k$means++ & 88.1 & 72.4 & 81.8 & 84.3 & 81.4 & 77.5 & 80.9 \\
        \midrule
        \multicolumn{8}{@{}l@{}}{CBMBR with various numbers of $k$means++ iterations} \\
        ~~niter=1 & 88.2 & 72.3 & 81.9 & 84.4 & 81.5 & 77.5 & 81.0 \\
        ~~niter=2 & 88.2 & 72.2 & 81.9 & 84.4 & 81.6 & 77.5 & 81.0 \\
        ~~niter=3 & 88.2 & 72.3 & 81.9 & 84.4 & 81.5 & 77.5 & 81.0 \\
        ~~niter=4 & 88.2 & 72.3 & 81.8 & 84.4 & 81.5 & 77.5 & 81.0 \\
        ~~niter=5 & 88.2 & 72.3 & 81.9 & 84.4 & 81.6 & 77.5 & 81.0 \\
        \midrule
        \multicolumn{8}{@{}l@{}}{CBMBR with various numbers of $k$means iterations} \\
        ~~niter=1 & 88.1 & 72.4 & 81.8 & 84.3 & 81.4 & 77.5 & 80.9 \\
        ~~niter=2 & 88.1 & 72.4 & 81.8 & 84.3 & 81.4 & 77.5 & 80.9 \\
        ~~niter=3 & 88.1 & 72.4 & 81.8 & 84.3 & 81.4 & 77.5 & 80.9 \\
        ~~niter=4 & 88.1 & 72.4 & 81.8 & 84.3 & 81.5 & 77.5 & 80.9 \\
        ~~niter=5 & 88.1 & 72.4 & 81.8 & 84.3 & 81.5 & 77.5 & 80.9 \\
        \midrule
        oracle & 89.9 & 76.3 & 84.0 & 87.3 & 84.2 & 80.2 & 83.7 \\
        \bottomrule
    \end{tabular}
    \caption{Translation quality (COMET\%) in the diverse translation candidates setting in the WMT'21 translation task.
    ``niter'' denotes the number of iterations of $k$means clustering.}
    \label{tab:results-mbr-valid}
\end{table}

\subsection{Decoding speed in the multi-system translation setting}

Table~\ref{tab:results-speed-multisystem} shows the decoding speed in the multi-system translation setting measure in the WMT'22 and WMT'23 En$\leftrightarrow$Ja translation tasks.
As shown in the table, our CBMBR improved the speed of the expected score calculation by 5.0 times compared with vanilla MBR and 1.5 times compared with PruneMBR in the multi-system setting.

\begin{table}[h]
    \centering
    \small
    \tabcolsep 3pt
    \begin{tabular}{@{}lrrrr@{}}
        \toprule
        Step & QE & MBR & PruneMBR & CBMBR \\
        \midrule
        Encode & \\ 
        ~~~hypotheses; $\mathcal{H}$ & -- & 198.1 & 198.7 & 199.1 \\
        ~~~source; $x$ & -- & 22.0 & 22.0 & 21.9 \\
        Rerank & 313.0 & -- & -- & -- \\
        Prune & -- & -- & 5.4 & -- \\
        $k$means++ & -- & -- & -- & 36.1 \\
        Utility function; $s$ & -- & 281.1 & 79.5 & 20.1 \\
        \midrule
        E2E & 336.0 & 511.9 & 306.0 & 278.4 \\
        \bottomrule
    \end{tabular}
    \caption{Average processing time per sentence (msec) in the multi-system translation setting measured in the WMT'22 and WMT'23 En$\leftrightarrow$Ja translation tasks.
    Note that ``E2E'' measures the end-to-end time, including miscellaneous processes.}
    \label{tab:results-speed-multisystem}
\end{table}

\subsection{Relationship between the number of centroids and translation quality in the diverse translation setting}
\begin{figure}
    \centering
    \includegraphics[width=\linewidth]{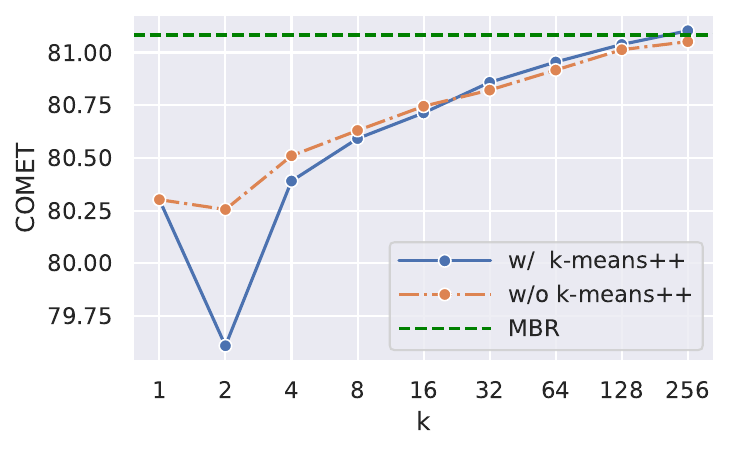}
    \caption{Translation quality of various $k$ in the diverse translation candidates setting.
    The scores are averaged COMET on WMT’21 En$\leftrightarrow$Ja, En$\leftrightarrow$De, and En$\leftrightarrow$Zh.
    }
    \label{fig:various-k-diverse}
\end{figure}

Figure~\ref{fig:various-k-diverse} shows the translation quality of various $k$ in the diverse translation candidates setting.
We observed that the COMET score increased as $k$ increased, except for an outlier, i.e., $k=2$.

\subsection{Evaluation of non-target metrics}
\begin{table}[t]
    \centering
    \small
    \tabcolsep 2.1pt
    \begin{tabular}{@{}lrrrrrrr@{}}
    \toprule
    Decoding & en-ja & ja-en & en-de & de-en & en-zh & zh-en & avg. \\
    \midrule
    MAP & 14.6 & \underline{10.9} & \textbf{25.1} & \textbf{27.3} & 28.2 & \textbf{16.1} & \textbf{20.4} \\
    QE & \underline{16.3} & \textbf{11.1} & 22.6 & 24.2 & 25.8 & 15.5 & 19.2 \\
    MBR & \textbf{16.4} & 10.6 & \underline{23.6} & \underline{25.8} & \textbf{28.3} & 15.6 & \underline{20.1} \\
    PruneMBR & \textbf{16.4} & 10.5 & \underline{23.6} & 25.7 & \underline{28.2} & 15.6 & 20.0 \\
    CBMBR & 16.2 & 9.8 & 23.1 & 25.4 & 27.6 & 15.5 & 19.6 \\
    ~~w/o $k$means++ & \textbf{16.4} & 10.4 & 23.5 & 25.7 & \textbf{28.3} & \underline{15.7} & 20.0 \\
    \midrule
    Oracle & 22.0 & 15.8 & 30.3 & 34.5 & 35.6 & 20.4 & 26.4 \\
    \bottomrule
    \end{tabular}
    \caption{
    BLEU scores~\cite{papineni-etal-2002-bleu} in the WMT'22 translation task with the diverse translation candidates setting.
    }
    \label{tab:results-wmt22-bleu}
\end{table}
\begin{table}[t]
    \centering
    \small
    \tabcolsep 2.1pt
    \begin{tabular}{@{}lrrrrrrr@{}}
    \toprule
    Decoding & en-ja & ja-en & en-de & de-en & en-zh & zh-en & avg. \\
    \midrule
    MAP & 56.6 & 54.3 & 65.2 & 66.9 & 61.2 & 56.8 & 60.2 \\
    QE & \textbf{63.7} & \textbf{60.1} & \textbf{70.0} & \textbf{69.6} & \textbf{65.8} & \textbf{62.2} & \textbf{65.2} \\
    MBR & \underline{62.7} & \underline{57.2} & \underline{68.5} & \underline{68.9} & \underline{65.1} & \underline{60.3} & \underline{63.8} \\
    PruneMBR & 62.6 & 57.1 & 68.4 & \underline{68.9} & \underline{65.1} & 60.2 & 63.7 \\
    CBMBR & 62.3 & 56.8 & 68.1 & 68.7 & 64.7 & 59.9 & 63.4 \\
    ~~w/o $k$means++ & 62.5 & 56.9 & 68.3 & 68.7 & 64.9 & 60.2 & 63.6 \\
    \midrule
    Oracle & 67.9 & 63.1 & 72.6 & 73.4 & 70.3 & 64.8 & 68.7 \\
    \bottomrule
    \end{tabular}
    \caption{
    BLEURT scores~\cite{sellam-etal-2020-bleurt} in the WMT'22 translation task with the diverse translation candidates setting.
    }
    \label{tab:results-wmt22-bleurt}
\end{table}

\begin{table}[t]
    \centering
    \small
    \begin{tabular}{@{}lrrrrr@{}}
        \toprule
        & \multicolumn{2}{c}{WMT'22} & \multicolumn{2}{c}{WMT'23} \\
        \cmidrule(lr){2-3} \cmidrule(lr){4-5}
        Decoding & en-ja & ja-en & en-ja & ja-en & avg. \\
        \midrule
        MAP & 24.2 & \textbf{23.1} & 20.7 & \textbf{21.9} & \textbf{22.5} \\
        QE & 23.3 & 20.7 & 20.0 & 19.9 & 21.0 \\
        MBR & \textbf{25.0} & \underline{21.9} & \textbf{21.6} & \underline{21.3} & \underline{22.4} \\
        PruneMBR & 23.8 & 21.7 & 20.5 & 21.2 & 21.8 \\
        CBMBR & 24.3 & 20.4 & 20.7 & 19.8 & 21.3 \\
        ~~w/o $k$means++ & \underline{24.8} & 21.8 & \underline{21.5} & 21.1 & 22.3 \\
        \midrule
        Oracle & 35.2 & 36.2 & 29.7 & 33.8 & 33.7 \\
    \bottomrule
    \end{tabular}
    \caption{BLEU scores~\cite{papineni-etal-2002-bleu}
    in the multi-system translation setting.
    }
    \label{tab:results-multisystem-bleu}
\end{table}

\begin{table}[t]
    \centering
    \small
    \begin{tabular}{@{}lrrrrr@{}}
        \toprule
        & \multicolumn{2}{c}{WMT'22} & \multicolumn{2}{c}{WMT'23} \\
        \cmidrule(lr){2-3} \cmidrule(lr){4-5}
        Decoding & en-ja & ja-en & en-ja & ja-en & avg. \\
        \midrule
        MAP & 65.1 & 67.0 & 59.6 & 67.1 & 64.7 \\
        QE & \textbf{68.8} & \textbf{68.6} & \textbf{63.6} & \textbf{68.9} & \textbf{67.5} \\
        MBR & \underline{68.3} & \underline{68.0} & \underline{63.5} & \underline{68.8} & \underline{67.2} \\
        PruneMBR & 66.5 & 66.9 & 61.3 & 67.5 & 65.6 \\
        CBMBR & \underline{68.3} & 67.4 & \underline{63.5} & 68.4 & 66.9 \\
        ~~w/o $k$means++ & \underline{68.3} & \underline{68.0} & \textbf{63.6} & 68.7 & 67.1 \\
        \midrule
        Oracle & 75.5 & 77.1 & 70.6 & 76.5 & 74.9 \\
        \bottomrule
    \end{tabular}
    \caption{BLEURT scores~\cite{sellam-etal-2020-bleurt}
    in the multi-system translation setting.
    }
    \label{tab:results-multisystem-bleurt}
\end{table}

We also evaluated translation quality using metrics that were not used as the utility function.
Table~\ref{tab:results-wmt22-bleu} and \ref{tab:results-wmt22-bleurt} show the results of the diverse translation candidates setting, and Table~\ref{tab:results-multisystem-bleu} and \ref{tab:results-multisystem-bleurt} show the results of the multi-system translation setting, respectively.

% \section{Additional Analysis}
% \subsection{Multimodality of translation}
\section{Multimodality of Translation Candidates}
\label{sec:analysis-multimodality}

\begin{table}[t]
    \centering
    \small
    \begin{tabular}{@{}lrrrrr@{}}
        \toprule
        & \multicolumn{2}{c}{WMT'22} & \multicolumn{2}{c}{WMT'23} \\
        \cmidrule(lr){2-3} \cmidrule(lr){4-5}
        Decoding & en-ja & ja-en & en-ja & ja-en & avg. \\
        \midrule
        MAP & 86.4 & 80.9 & 83.5 & 80.4 & 82.8 \\
        QE & 89.8 & 82.6 & 87.6 & 82.3 & 85.6 \\
        MBR & \underline{90.5} & \textbf{84.1} & 88.7 & \underline{83.7} & 86.7 \\
        PruneMBR & 88.9 & 82.8 & 86.6 & 82.2 & 85.1 \\
        CBMBR & \textbf{90.9} & \textbf{84.1} & \textbf{89.2} & \textbf{83.8} & \textbf{87.0} \\
        ~~w/o $k$means++ & \underline{90.5} & \textbf{84.1} & \underline{88.8} & \underline{83.7} & \underline{86.8} \\
        CBMBR$_\text{cnt}$  & 90.4 & 83.9 & 88.6 & 83.5 & 86.6 \\
        ~~w/o $k$means++ & 90.4 & \underline{84.0} & 88.6 & 83.6 & 86.6 \\
        \midrule
        Oracle & 93.4 & 89.4 & 91.9 & 88.5 & 90.8 \\
        \bottomrule
    \end{tabular}
    \caption{Results of the multi-system translation setting with weighting by the numbers of samples.}
    \label{tab:result-count-weight}
\end{table}

In the multi-system translation setting, CBMBR also outperformed vanilla MBR in terms of translation quality, as shown in Table~\ref{tab:results-biased} and Figure~\ref{fig:cbmbr-various-k}.
In this section, we discuss the multimodal nature of translation, two approximations of MBR decoding, and why our CBMBR outperformed vanilla MBR in the multi-system translation setting.

The $n$-best translations generated by beam search are often similar to each other~\citep{vijayakumar-etal-2018-diverse}.
To diversify the candidates while maximizing translation quality, \citet{deguchi-etal-2023-naist} generated the 50-best translation sets from each translation system, which resulted in candidates that exhibited multimodality.
Vanilla MBR decoding calculates the expected score by treating all samples equally, which means that it is prone to being affected by the number of similar translation samples for the candidates that have such a multimodal distribution.

Now, there are two approximation variants of MBR decoding in our CBMBR.
One is our proposed method, which calculates the expected score using centroid representations:
\begin{equation}
    y^\ast_\text{CBMBR} = \argmax_{h \in \mathcal{H}} \expect_{\bm{c} \in \mathcal{C}} \left[ s(f(x), f(h), \bm{c}) \right].
\end{equation}
The other, CBMBR$_\text{cnt}$, multiplies each centroid-based score by the weight according to the number of samples in each cluster:
\begin{align}
    &y^\ast_{\text{CBMBR}_\text{cnt}} = \nonumber \\
    &\argmax_{h \in \mathcal{H}} \expect_{\bm{c} \in \mathcal{C}} \left[ s(f(x), f(h), \bm{c}) \times w(\bm{c}) \right],
    \label{eq:CBMBR-cnt}
\end{align}
where $w\colon \mathbb{R}^D \to [0,1]$ returns the weight of the given centroid as follows:
\begin{align}
    &w(\bm{c}) = \frac{\mathrm{count}(\bm{c})}{\sum_{i=1}^k \mathrm{count}(\bm{c}_i)}, \\
    &\mathrm{count}(\bm{c}) = \left| \left\{\, \hat{y} \in \hat{\mathcal{Y}} : \bm{c} = \mathrm{NN}\left( f(\hat{y}), \mathcal{C} \right) \,\right\} \right|, \\
    &\mathrm{NN}(\bm{q}, \mathcal{C}) = \argmin_{\bm{c} \in \mathcal{C}} \lVert \bm{q} - \bm{c} \rVert_2,
\end{align}
where $\mathrm{NN}\colon \mathbb{R}^D \times \mathcal{C} \to \mathbb{R}^D$ finds the nearest neighbor centroid of a vector $\bm{q} \in \mathbb{R}^D$ from the given set of centroids $\mathcal{C}$ and $\mathrm{count}\colon \mathbb{R}^D \to \mathbb{N} \cup \{ 0 \}$ counts the number of samples in the cluster of the given centroid.
CBMBR${}_\text{cnt}$, which uses the number of samples in a cluster, can be regarded as more accurately approximating vanilla MBR compared with our CBMBR, which ignores the number of samples in a cluster.

We compared the translation quality of our CBMBR and CBMBR$_\text{cnt}$ with that in the multi-system translation setting.
Table~\ref{tab:result-count-weight} shows the results.
From the results, the difference between vanilla MBR and CBMBR$_\text{cnt}$ narrowed to 0.1\% and degraded by 0.6\% compared with CBMBR; that is, CBMBR$_\text{cnt}$, which more accurately approximates vanilla MBR, was worse than our proposed CBMBR.
We attribute this observation to the biased distribution caused by beam search or sampling.
With CBMBR, we can robustly decode against bias.

\end{document}